\documentclass[10pt,twocolumn,letterpaper]{article}

\usepackage{cvpr}
\usepackage{algorithm}
\usepackage{algpseudocode}

\usepackage{amsmath,amsfonts,bm}
\usepackage{multirow}
\usepackage{dblfloatfix}
\usepackage{breqn}

\def\eqref#1{equation~\ref{#1}}

\def\1{\bm{1}}

\DeclareMathAlphabet{\mathsfit}{\encodingdefault}{\sfdefault}{m}{sl}
\SetMathAlphabet{\mathsfit}{bold}{\encodingdefault}{\sfdefault}{bx}{n}

\definecolor{cvprblue}{rgb}{0.21,0.49,0.74}
\usepackage[pagebackref,breaklinks,colorlinks,allcolors=cvprblue]{hyperref}

\def\paperID{****} 
\def\confName{CVPR}
\def\confYear{2025}

\title{\ours: Learning a Unified Constitutive Model for Inverse Physics Simulation}

\author{Himangi Mittal$^{1}$ ~~~~~~~~ Peiye Zhuang$^2$ ~~~~~~~~ Hsin-Ying Lee$^2$ ~~~~~~~~ Shubham Tulsiani$^1$ \\
$^1$Carnegie Mellon University ~~~~ $^2$Snap Inc. \\
}

\newcommand{\Dcal}{\mathcal{D}}
\newcommand{\inpx}{\mathbf{x}}
\newcommand{\inpv}{\mathbf{v}}
\newcommand{\inpC}{\mathbf{C}}
\newcommand{\inpF}{\mathbf{F}}
\newcommand{\inpS}{\mathbf{S}}
\newcommand{\ours}{UniPhy\xspace}

\begin{document}
\twocolumn[{%
\renewcommand\twocolumn[1][]{#1}%
\maketitle
\begin{center}
\vspace{-2em}
\href{https://himangim.github.io/UniPhy}{\texttt{https://himangim.github.io/UniPhy}}
\end{center}
\begin{center}
    \centering
    \captionsetup{type=figure}
    \includegraphics[trim=0cm 1.5cm 0cm 0cm, clip, width=\linewidth]{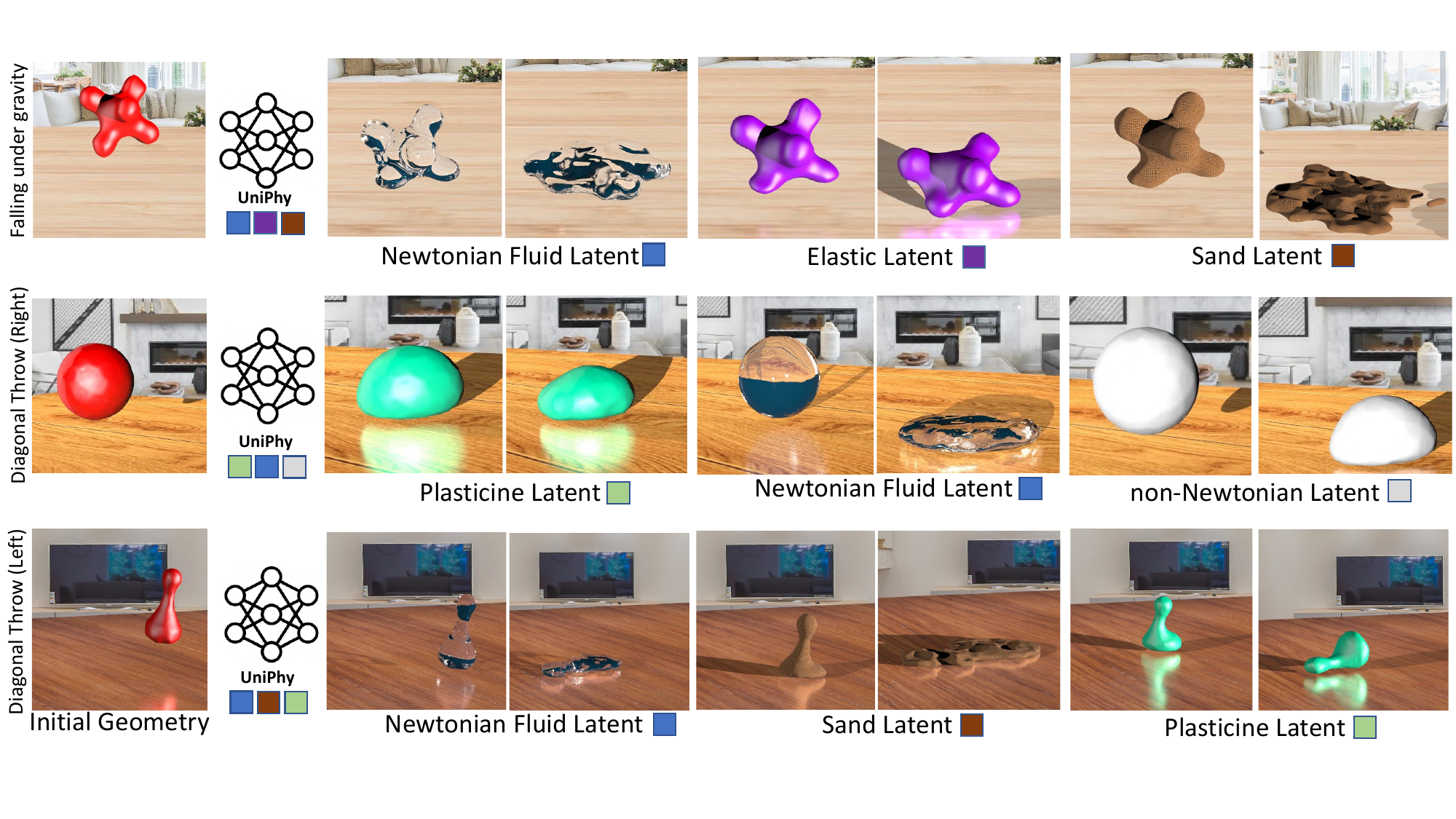}
    \caption{We present \textbf{UniPhy}, a unified latent-conditioned neural model which learns a common latent space to encode the properties of diverse materials. At inference, given motion observations for a system with unknown material parameters, \ours\ allows material inference via differentiable simulation-based latent optimization. These inferred material latents 
    can be used to simulate new trajectories that reflect the behavior of the underlying material. For example, in the first row, given the initial geometry of a toy and an optimized latent representing newtonian fluid~(\textcolor{blue}{blue block}), the geometry spreads when it hits the floor, whereas for an optimized latent representing elastic materials~(\textcolor{violet}{purple block}), it squeezes.}
    \label{fig:teaser}
\end{center}%
}]

\begin{abstract}
We propose \ours, a common latent-conditioned neural constitutive model that can encode the physical properties of diverse materials. At inference \ours\ allows `inverse simulation' \ie inferring material properties by  optimizing the scene-specific latent to match the available observations via differentiable simulation. In contrast to existing methods that treat such inference as system identification, \ours\ does not rely on user-specified material type information. Compared to prior neural constitutive modeling approaches which learn instance specific networks, the shared training across materials improves both, robustness and accuracy of the estimates. We train \ours\ using simulated trajectories across diverse geometries and materials -- elastic, plasticine, sand, and fluids~(Newtonian \& non-Newtonian). At inference, given an object with unknown material properties, \ours\ can infer the material properties via latent optimization to match the motion observations, and can then allow re-simulating the object under diverse scenarios. We compare \ours\ against prior inverse simulation methods, and show that the inference from \ours\ enables more accurate replay and re-simulation under novel conditions.

\end{abstract}

\section{Introduction}
\label{sec:intro}

The rich history of data-driven visual learning has offered us a common lesson across tasks -- unified expressive models can lead to improved performance and broader adaptation than their constrained counterparts. For example, initial detection methods~\cite{Girshick_2015_CVPR} relied on category-level templates but recent widely-adopted systems~\cite{carion2020end, kirillov2023segment} learn a common model across generic objects. Similarly, initial learning-based methods for 3D object reconstruction~\cite{kanazawa2018learning} learned category-level deformable models, but recent approaches learn category-agnostic reconstruction systems. 
Unified systems such as these have led to a remarkable progress across tasks related to semantic and spatial understanding from perceptual input, and in this work, we seek to take a similar (albeit small) step for physical understanding.

Akin to the progress in differentiable rendering~\cite{Liu_2019_ICCV,mildenhall2021nerf,kerbl20233d} that enabled `inverse rendering' \ie optimizing 3D representations given multi-view captures, there has been impressive progress in developing efficient and expressive differentiable physics simulators~\cite{hu2019difftaichi}. Leveraging these can similarly enable `inverse simulation', where we can infer the physical properties of the underlying system given observations about its evolution over time. More specifically, we consider scenarios where the `material properties' \ie how it behaves under deformation (captured by `constitutive functions' that predict internal forces given deformation) are unknown and need to be estimated via inverse simulation. Prior methods~\cite{li2023pac} often treat this task as a `system identification' problem, where user specifies a parametric constitutive model (\eg by specifying whether the material is Elastic or a Newtonian fluid), and some unknown continuous parameters (\eg Young's modulus) are optimized via differentiable simulation. Such material-type dependent approaches, however, require user input and do not present a unified solution across generic objects. Closer to our framework, Ma \etal~\cite{ma2023learning} presented a framework for learning a neural constitutive model for capturing generic material, but required a \emph{separate} network for each scene.

In this work, we build upon this insight that neural models can be learned to accurately capture the physical constitutive models across materials and learn \ours: a \emph{unified} neural constitutive model that can capture diverse materials (\eg plasticine materials as well as fluids) and instantiations (\eg varying viscosity). In particular, \ours\ models the variations across physical materials via a common latent-conditioned neural model, where different materials correspond to different latents that condition a common neural constitutive model -- the material properties are thus captured by a corresponding latent representation. Unlike NCLaw~\cite{ma2023learning}, this allows using a common network across simulations and unlike system identification methods, this obviates the need for user-specified type-dependent  (parametric) constitutive models.

We integrate \ours\ with a differentiable MPM~\cite{bardenhagen2000material} simulator, and train it using several simulated trajectories of systems with varying physical properties. We show that this single model is effectively able to capture the variation across diverse materials, and that it can enable inverse simulation given just 3D (or 2D) motion of the system by simply optimizing the corresponding latent. We compare our inference to prior methods that also seek to infer physical properties given 3D trajectories, and show that \ours\ yields clear improvements -- accurately allowing simulated replay as well as re-simulation under different initial conditions.

\section{Related Works}

\begin{figure*}
  \centering
  \captionsetup{type=figure}
  \includegraphics[trim=0cm 7cm 0cm 0cm, clip,width=\textwidth]{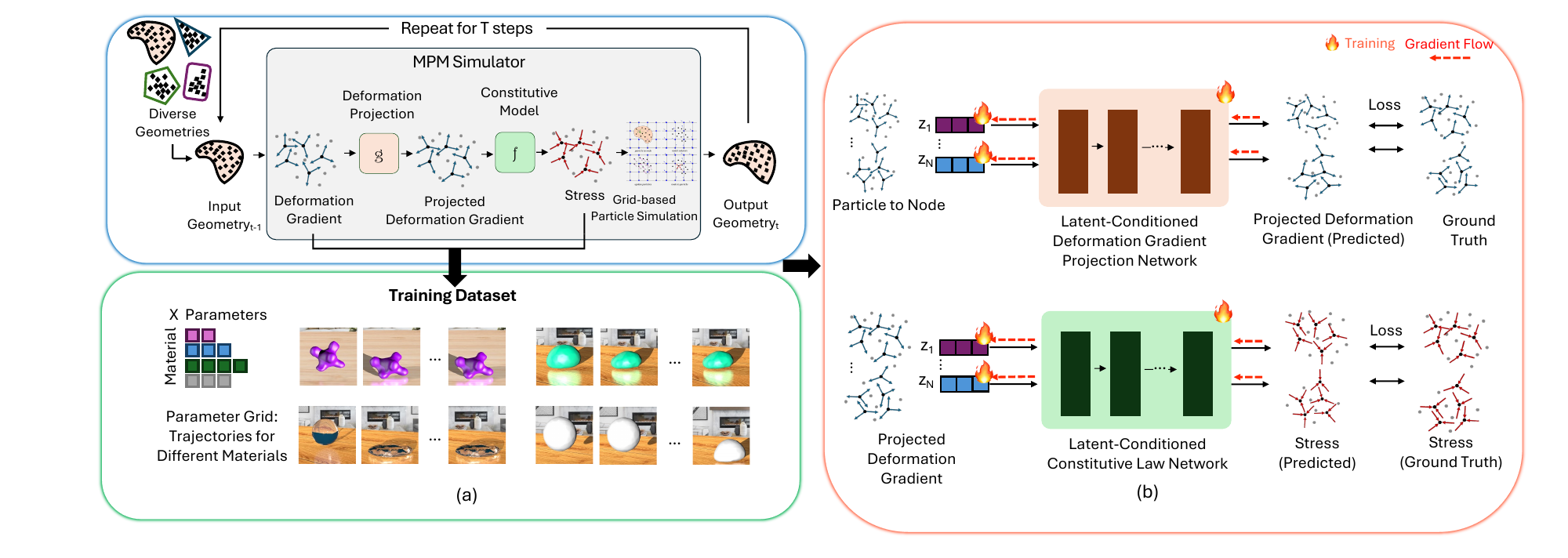}
  \vspace{-2em}
  \caption{\textbf{Overview} (a)~We use the Material Point Method simulator to generate a dataset with various geometries, motions, and material parameters~(\eg Young's modulus) as represented in the parameter grid. The dataset comprises of the particle to node deformation gradient matrix, projected deformation gradient matrix, and stress matrix. The visualizations show some example trajectories of different materials. (b)~Next, we use the dataset to train the latent-conditioned deformation gradient projection network~($g_\phi$) in \textcolor{orange}{orange} with the particle-to-node deformation gradient matrix and the latent as input and the latent-conditioned constitutive law network~($f_\theta$) in \textcolor{green}{green} block with the projected deformation gradient matrix and the latent as the input. Each trajectory in the dataset is initialized with its unique random latent. Note that the same trajectory latent is given to both the networks. We use the $\mathcal{L}2$ objective function between the predicted and ground truth deformation gradient matrix, and the predicted and ground truth stress matrix to optimize both the neural networks and the latent.} 
  \label{fig:stage_1}
  \vspace{-1em}
\end{figure*}

\noindent \textbf{Constitutive Laws and Neural Constitutive Models.} A constitutive law defines the relationship between strain and stress and measures the stress response of the material under deformation. These laws have been defined for various materials such as elastic~\cite{treloar1943elasticity,fung1967elasticity,arruda1993three}, sand~\cite{klar2016drucker}, viscoelastic and elastoplastic solids~\cite{fang2019silly}, foam~\cite{yue2015continuum}, fluids~\cite{chhabra2006bubbles}, and plasticity~\cite{mises1913mechanik}. Recent methods~\cite{liu2022learning,klein2022polyconvex,tartakovsky2018learning,wang2018multiscale,vlassis2020geometric,vlassis2021sobolev,vlassis2022component,vlassis2023geometric,vlassis2022molecular,ma2023learning,han2024learning,li2022plasticitynet} approximate these constitutive laws through neural networks due to their generalizable and robust nature. Our work explores the same domain of approximating the constitutive law as well as the deformation gradient projection mapping through neural networks, but unlike prior approaches, we seek to learn a unified model across materials.

\noindent \textbf{Inference via Differentiable Simulation.} Implementing the PDE simulations in a differentiable manner is helpful to perform gradient-based optimization and backpropagate the gradients across the simulation states~\cite{hahn2019real2sim,liang2019differentiable,ma2021diffaqua,hu2019chainqueen,huang2021plasticinelab,du2021diffpd,qiao2021differentiable,du2020functional,de2018end,geilinger2020add,xian2023fluidlab,degrave2019differentiable}. Previous methods use differentiable simulations to estimate the constitutive parameters~\cite{ma2022risp,chen2022virtual}, perform system identification~\cite{li2023pac} or approximate the constitutive models. However, these works are limited to simpler settings such as limiting the formulation to elastic materials~\cite{wang2020learning,huang2020learning} or training separate networks for each materials~\cite{ma2023learning}. Our work leverages differentiable Material Point Method~(MPM) to develop a single, unified model for learning across diverse material properties.

\vspace{1mm}
\noindent \textbf{3D Dynamics with Physics.} Recent approaches such as dynamic NeRF~\cite{li2023dynibar,park2021hypernerf,pumarola2021d,guo2023forward} and dynamic 3D Gaussians~\cite{Huang_2024_CVPR,luiten2024dynamic} extend geometric reasoning to a dynamic domain by first learning the scene in a canonical space and then mapping this space into a deformed at a particular timestep. While they develop a dynamic 3D geometric understanding of the scene at each timestep, their representations do not capture the underlying physics (for example, internal and external forces) and material-related behaviors that come into play and influence the trajectory of the objects in the scene. While some recent approaches~\cite{xie2024physgaussian,zhang2025physdreamer} show that these 3D (gaussian-based) representations can be physically grounded and simulated, they rely on user-specified material properties and cannot infer these from observations. Closer to our work, PAC-NeRF~\cite{li2023pac} combined deformable NeRFs with differentiable simulation to infer material properties from visual input, but treated this as a system identification task with known material type. We hope that a unified system such as ours will make it easier to incorporate physical reasoning for such approaches by additionally incorporating (latent) material parametrization in the representation.

\noindent \textbf{Neural Networks for Physics.} The integration of machine learning with physics-based simulations has been gaining significant interest. In addition to the neural network based approaches, graph-based networks have also been explored to approximate the PDE solutions for differentiable simulation~\cite{belbute2020combining} and to simulate various physical domains by considering the particle state of the system as node and computing the dynamics via message-passing~\cite{sanchez2020learning,pfaff2020learning}, and PINN~\cite{raissi2019physics,karniadakis2021physics} which uses the residual of partial differential equation in the loss function to learn physics priors. Although learning via these methods seems to be a promising direction, these approaches seek to learn the entire simulation pipeline through their network. Whereas, in our work, we incorporate learned modules in classical simulators that can be applied to various scenarios and can lead to a more broadly applicable system.

\section{Learning a Unified Neural Material Model}

During inference, given an object trajectory with unknown material properties, our aim is to infer the material properties of the object from the positions of the particles. We achieve this by first learning neural representations of various materials in a latent space and then searching for a latent in the latent space during inference such that this latent represents the material properties of the given object. Learning the material properties via latent space helps us to simulate an object trajectory during inference without having to make any assumptions such as knowing the material properties beforehand~\cite{li2023pac} or having to train a separate model for each material~\cite{ma2023learning}.

\begin{figure*}
  \centering
  \captionsetup{type=figure}
  \vspace{-2em}
  \includegraphics[trim=0cm 12cm 7cm 0cm, clip,width=\textwidth]{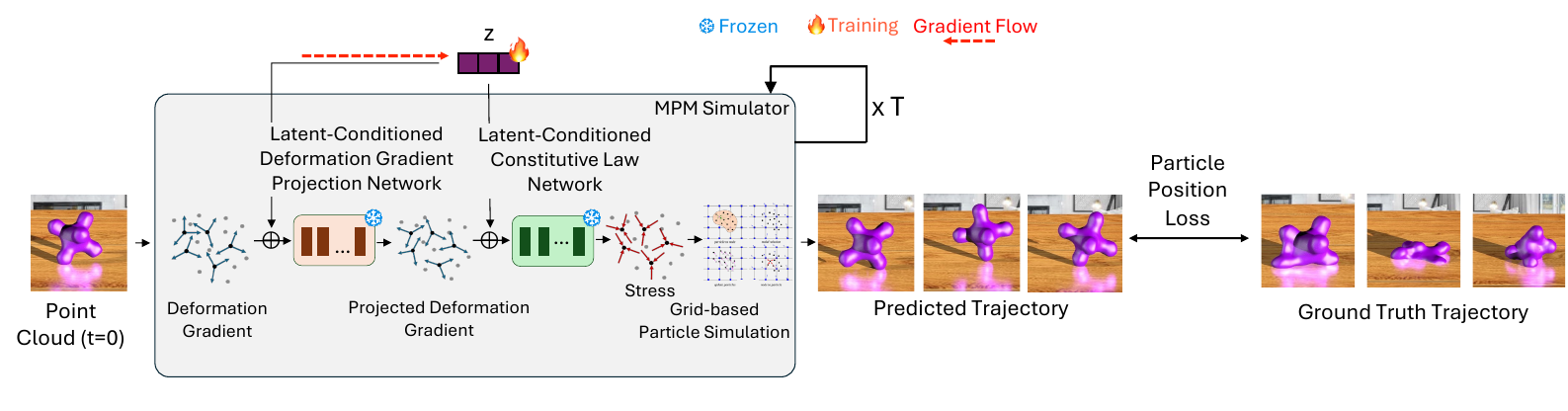}
  \caption{We show the inference setup in the figure where given an object trajectory along with a random latent, we do not have the material properties beforehand. After training the deformation gradient projection function network~($g_\phi$) and constitutive law network~($f_\theta$), we embed the networks~(weights frozen) in the differentiable Material Point Method~(MPM) simulator and optimize the latent~($z$) using the $\mathcal{L}2$ loss between the predicted trajectory and ground truth trajectory. The latent after optimization is able to capture the material properties of the given object.}  
  \label{fig:stage_2}
  \vspace{-1em}
\end{figure*}

To learn these neural representations of the materials, we propose a single, unified network system conditioned on a latent vector to model the behavior of the different materials. For our approach, we use the Material Point Method~(MPM) as the differentiable simulation method discussed in Section~\ref{sec:mpm}. We then describe the training and inference pipeline to train our network conditioned on the latent~(Section~\ref{sec:latent}). Figure~\ref{fig:stage_1} and \ref{fig:stage_2} provide a full overview of \ours.

\subsection{Background}
\label{sec:mpm}

In this section, we briefly describe the Material Point Method~(MPM)~(as can be seen in Figure~\ref{fig:stage_1}a). MPM is an effective hybrid particle/grid method for simulating various kinds of materials. In MPM, we represent the state of a simulation trajectory as follows: $\inpx \in \mathbb{R}^{P \times 3}$ as the position, $\inpv \in \mathbb{R}^{P \times 3}$ as the velocity, $\inpC \in \mathbb{R}^{P \times 3 \times 3}$ as the affine velocity, and $\inpF \in \mathbb{R}^{P \times 3 \times 3}$ as the deformation gradient. The affine velocity~($\inpC$) represents the local velocity gradient field of a point. Each particle has a mass $\mathbf{m} \in \mathbb{R}$.

The first step is to compute the deformation gradient and internal forces in the Lagrangian space. The deformation gradient is passed through a projection mapping function to get a projected deformation gradient and then internal forces are computed defined using the projected deformation gradient. The next step is to transfer the particles' information to the grid using the Affine Particle-in-Cell~(APIC)~\cite{jiang2015affine} method. The mass and momentum of each particle is distributed to its neighboring grid nodes on the basis of a weighting mechanism. Explicit grid forces, laws of conservation, and boundary conditions are computed on the node to get the final nodal mass and velocity. Finally, the particles position, velocity, and affine velocity are updated using grid to particle transfer.

\textbf{Deformation Gradient Projection Function:} The deformation gradient represents the deformation of an object locally. 
Each material has its own deformation gradient projection mapping function that defines the local deformation behavior. For a particle indexed by $p$ and the deformation gradient of the particle defined by $\mathbf{F}^{p,t}$ at timestep $t$, the projection function $\phi$ of the deformation gradient can be defined as,
\begin{equation}
    \mathbf{F}_{proj}^{p, t} = \phi_{mat=\{k_1,..k_n\}} (\mathbf{F}^{p,t})
\end{equation}
This projection function is dependent on the material parameters $mat={\{k_1,..k_n\}}$. For example, the projection function $\phi$ for elastic material is an identity function whereas for plasticine materials, the projection function is used to pull the deformation gradient back onto the yield surface and is dependent on parameters such as yield stress and Lam\'e parameters. 

\textbf{Constitutive Model:} Once we have the projected deformation gradient, it is used as a strain to compute the stress/internal forces experienced by an object through a constitutive model. A constitutive model $\psi$ defines the relationship between the strain and stress forces $\inpS^{p,t}$ for a particle indexed by $p$ at a timestep $t$ as, 
\begin{equation}
    \inpS^{p,t} = \psi_{mat=\{k_1,..k_n\}} (\mathbf{F}_{proj}^{p, t})
\end{equation}
The constitutive model is also a material-dependent function that defines the different stress behaviors of the various materials and is dependent on the material parameters such Lam\'e parameters, friction angle, yield stress, \etc. Please refer to the appendix for the projection function and constitutive models of different materials.

Once the stress forces are calculated, they are used to update the momentum and velocity of the grid node. Finally, the updated information of grid velocity and momentum is transferred back to the particles using the same APIC method and the particles are advected with the new velocities to new positions: $\inpx^{i, t+1} = \inpx^{i, t} + \Delta t \inpv^{i,t+1}$.

\subsection{Latent-Conditioned Unified Material Model}
\label{sec:latent}

In the above simulation approach, the deformation gradient projection function and constitutive model are dependent on the assumption that the material properties are given beforehand. Such an assumption is not feasible in the real-world where it can be difficult to know the material properties in advance. To address this, we introduce \ours, a unified, neural model that can learn representations across diverse materials without the need for human-in-the-loop to define material information beforehand. We achieve this by learning neural representations of materials through a latent space $\mathcal{Z}$. At test time, \ours can then be used to infer material properties of an object directly from particle positions.

To embed material information in the latent, we use a learnable latent $\mathbf{z}$ to train two neural networks jointly -- latent-conditioned deformation gradient projection network~($g_\phi$)~(Section~\ref{sec:latent-F}) and latent-conditioned neural constitutive law network~($f_\theta$)~(Section~\ref{sec:latent-stress}). We map the latent vector to the deformation gradient projection function and constitutive law of a material. By conditioning the neural networks on the latent, we can use a single, unified network to model the behavior of different materials. We train the two latent-conditioned networks by collecting a wide range of simulations from the Material Point Method simulator which gives us a dataset $\Dcal=\left[\{\{(\mathbf{F}^{p,t}_n, \mathbf{F}^{p, t}_{proj, n} , \mathbf{S}^{p,t}_{n})_{n=1}^{N}\}_{p=1}^{P}\}_{t=1}^{T}\right]$ containing $N$ trajectories with $P$ particles and $T$ timesteps~(Figure~\ref{fig:stage_1}a). 

\subsection{Latent-Conditioned Deformation Gradient Projection Network}
\label{sec:latent-F}

Given an object trajectory consisting of $\displaystyle P$ particles and comprising of $\displaystyle T$ timesteps, each particle at a timestep has a deformation gradient $\mathbf{F} \in \mathbb{R}^{3 \times 3}$  which defines the local deformation around the particle as compared to the initial state. The projection of the deformation gradient is used to pull it back onto the elastic region/yield surface giving us $\mathbf{F}_{proj} \in \mathbb{R}^{3 \times 3}$. 

To embed the material-dependent deformation gradient projection function in the latent space, we collect $\displaystyle N$ number of trajectories in our dataset $\Dcal$. Each trajectory gives us $\displaystyle P \times \displaystyle T$ pairs of ($\mathbf{F}^{p,t}$, $\mathbf{F}^{p,t}_{proj}$)~(Figure~\ref{fig:stage_1}a). For an object trajectory indexed by ${n}$, we initialize each trajectory with a latent vector $\mathbf{z}_n$ such that its corresponding pairs of ($\mathbf{F}^{p,t}$, $\mathbf{F}^{p,t}_{proj}$) are associated with the latent $\mathbf{z}_n$. Then, we train the neural network $g_{\phi}$, conditioned on the latent $\mathbf{z}_n$, to approximate the projection function of the object's material~(Figure~\ref{fig:stage_1}b),
\begin{equation}
    \mathbf{\hat{F}}_{proj, n}^{p, t} = g_\phi(\mathbf{F}^{p, t}_{n}, \mathbf{z}_n)
\end{equation}

The parameter $\phi$ learns the deformation gradient projection of a material and when jointly trained with a latent, the latent space embeds the projection function of all the materials. We train this latent-conditioned deformation gradient projection network jointly with another neural network that approximates the constitutive law of the material which we discuss in the following section~(as seen in Figure~\ref{fig:stage_1}b).

\subsection{Latent-Conditioned Constitutive Law Network}  
\label{sec:latent-stress}

For the same $\displaystyle N$ number of trajectories~(used in Section~\ref{sec:latent-F}), each trajectory also gives us $\displaystyle P \times \displaystyle T$ pairs of ($\mathbf{F}^{p,t}_{proj}, \mathbf{S}^{p,t}$)~(Figure~\ref{fig:stage_1}a). We give the projected deformation gradient $\mathbf{F}^{p,t}_{proj}$ for a particle $p$ at a timestep $t$ as input to the latent-conditioned neural constitutive law network $f_{\theta}$ to predict the stress $\mathbf{S}^{p,t} \in \mathbb{R}^{3 \times 3}$ experienced by the particle at that timestep.

 For a pair of ($\mathbf{F}^{p,t}_{proj}, \mathbf{S}^{p,t}$) associated with an object trajectory indexed by ${n}$, we condition the neural network $f_{\theta}$ on the same latent $\mathbf{z}_n$ used to condition $g_\phi$ to predict stress as~(Figure~\ref{fig:stage_1}b),
\begin{equation}
    \mathbf{\hat{S}}^{p,t}_{n} = f_\theta (\mathbf{F}_{proj,n}^{p,t}, \mathbf{z}_n)
    \label{eq:pos_loss}
\end{equation}

The parameter $\theta$ approximates the constitutive law of a material and by conditioning the network on the latent, we are able to embed the properties and behavior of all the materials in a common latent space.

To train the latent-conditioned neural networks $g_\phi$ and $f_\theta$ jointly with the latent vector $\mathbf{z}$, we use two $\mathcal{L}2$ objective functions: 1) \textbf{Deformation gradient projection function loss:} Loss between the predicted and ground truth deformation gradient projection values, $\mathbf{\hat{F}}^{p,t}_{proj,n}$ and $\mathbf{F}^{p,t}_{proj,n}$, respectively, 2) \textbf{Stress Constitutive Law loss:} Loss between the predicted stress values $\mathbf{\hat{S}}^{p,t}_{n}$ and ground truth $\mathbf{S}^{p,t}_{n}$ stress. The loss functions are defined as follows,
\vspace{-1em}
\begin{dmath}
    \underset{\theta,\phi, \{{z_n}\}}{\min } \sum_n \sum_t \sum_p \left(\mathcal{L} (\mathbf{\hat{F}}^{p,t}_{proj,n}, \mathbf{F}^{p,t}_{proj,n}) + \mathcal{L}(\mathbf{\hat{S}}^{p,t}_{n}, \mathbf{S}^{p,t}_{n}) + \frac{1}{\sigma^2}\Vert \mathbf{z}_n \Vert^2 \right)
    \label{eq:fproj_stress_loss}
\end{dmath}

We backpropagate the gradients of the loss function in Equation~\ref{eq:fproj_stress_loss} to the latent-conditioned neural networks $g_\phi$ and $f_\theta$ to jointly optimize each of the network weights $\{\phi, \theta\}$ and the latent vector $\mathbf{z}$. Through this training, we encode the material properties of deformation gradient projection function and constitutive law in the latent $\mathbf{z}_n$ and learn latent representations to model different materials.

\section{Inferring Material Properties via Differentiable Simulation}

During inference, for a given object trajectory, we assume that we are not aware of its material properties and we only know the initial state~(object geometry) as well as the future states of the object. Our learned model \ours enables us capture the material properties through a latent vector corresponding to that trajectory which instantiates the deformation gradient and constitutive law of the material of the object. To identify the material properties, we optimize this latent and alleviate to have a human-in-the-loop to define the material information. As an application, this method allows us to infer the material properties of an object via differentiable simulation-based latent optimization.

We use the optimized $f_\theta$, $g_\phi$ networks and randomly initialize the latent vector $\mathbf{z}$ from $\mathcal{N} (0, \mathbf{I})$. We freeze the parameters $\theta$ and $\phi$, and only optimize the latent vector $\mathbf{z}$~(Figure~\ref{fig:stage_2}). To optimize the latent to capture the material properties of the object during inference, we leverage the particle positions of the object trajectory as a signal for the latent to learn the material properties. For example, the trajectory motion of a bouncing elastic ball would give a different signal to the latent when compared to the trajectory motion of a sand ball which is hitting and spreading on the floor. 

In this stage, we embed our latent-conditioned networks inside the differentiable Material Point Method simulator to optimize the latent~(Figure~\ref{fig:stage_2}). Given a trajectory and a randomly initialized latent $\mathbf{z}$, we use our projection neural network $g_\phi$ to predict the deformation gradient projected values and the constitutive law neural network $f_\theta$ to predict the stress values. We use these predicted values in the (differentiable) MPM simulator to obtain the simulated particle positions trajectory $\{\hat{x}\}$. We optimize the latent $\mathbf{z}$ using a loss function $\mathcal{L}$ that computes the difference between the observed and simulated positions of each particle over time: 

\begin{align}
\hat{x} = MPM(g_\phi(., \mathbf{z}), f_\theta (., \mathbf{z})) \\
\mathbf{\hat{z}} = \underset{\mathbf{{z}}}{\min}~ \mathcal{L}\left(\{\hat{x}\}, \{{x}^{gt}\} \right)
\label{eq:pos_loss_inference}
\end{align}

The optimized latent succinctly captures the underlying material properties of the object, and can be deployed for simulating diverse trajectories of this object/material with different velocities and starting positions.

\section{Experiments}

\paragraph{Dataset.}~We train and evaluate \ours on five kinds of materials -- elastic, plasticine, Newtonian fluids, non-Newtonian fluids, and sand. Our dataset includes objects with various geometries, with motion in different directions of falling under gravity, horizontal rolling motion, and throwing an object diagonally, and having a diverse range of material physical parameters. In all, we create 200 trajectories per material. Please find details for the constitutive laws and physical parameters of the materials in the Appendix. 

\vspace{-1.2em}
\paragraph{Baseline and Ablations.} We compare our method with NCLaw~\cite{ma2023learning} that presents a hybrid NN-PDE approach for learning PDE dynamics, spline~\cite{xu2015nonlinear} which approximates the elastic constitutive laws using Bezier splines, neural~\cite{wang2020learning} that learns the constitutive laws via neural network, and gnn~\cite{sanchez2020learning} which learns to simulate via graph neural nets.

\vspace{-1.2em}
\paragraph{Implementation Details.} The latent-conditioned constitutive law network, $f_{\theta}$, comprises 5 linear layers with 128 as the hidden dimension and 32 as the dimension of the latent. Additionally, the latent-conditioned deformation gradient projection network, $g_{\phi}$ consists of 5 linear layers with a hidden dimension of 32. We train the network parameters and the latent with a learning rate of 1e-3 and 1e-2 respectively using the AdamW optimizer during training. For training, we initialize the material latent code randomly from $\mathcal{N}(0., 1.)$ and concatenate the network input with the latent. During inference, we optimize the latent with a learning rate of 1e-3 using the AdamW optimizer. We initialize the latent during inference from the best performing cluster centers found using K-means clustering on the learned latent space. We use the same implementation setup for all the materials.

\begin{table}[h]
\small
\resizebox{\linewidth}{!}{%
\centering
\begin{tabular}{p{2.1cm}||p{0.8cm}p{0.8cm}p{0.8cm}p{0.8cm}}
\hline
\multicolumn{1}{c||}{\textbf{Method}} & \multicolumn{1}{c}{\textbf{Elastic}} & \multicolumn{1}{c}{\textbf{Sand}} & \multicolumn{1}{c}{\textbf{Plasticine}} & \multicolumn{1}{c}{\textbf{Newtonian}} \\ \hline

\textbf{spline}~\cite{xu2015nonlinear} & \multicolumn{1}{c}{2.4e-1} & \multicolumn{1}{c}{3.2e-1} & \multicolumn{1}{c}{3.0e-1} & \multicolumn{1}{c}{3.2e-1} \\
\textbf{neural}~\cite{wang2020learning} & \multicolumn{1}{c}{1.2e-5} & \multicolumn{1}{c}{1.2e-2} & \multicolumn{1}{c}{1.9e-1} & \multicolumn{1}{c}{2.9e-2} \\
\textbf{gnn}~\cite{sanchez2020learning} & \multicolumn{1}{c}{2.1e-2} & \multicolumn{1}{c}{1.1e-2} & \multicolumn{1}{c}{8.7e-3} & \multicolumn{1}{c}{3.2e-2} \\
\textbf{nclaw}~\cite{ma2023learning} &  \multicolumn{1}{c}{2.4e-4} & \multicolumn{1}{c}{2.6e-5} & \multicolumn{1}{c}{6.5e-5} & \multicolumn{1}{c}{2.0e-5} \\
\textbf{ours}   & \multicolumn{1}{c}{\textbf{5.2e-6}}   &   \multicolumn{1}{c}{\textbf{1.5e-5}}  & \multicolumn{1}{c}{\textbf{3.9e-5}}   & \multicolumn{1}{c}{\textbf{1.1e-6}} \\ 
\hline 
\textbf{nclaw~(w/o TF)} & \multicolumn{1}{c}{3.6e-3} & \multicolumn{1}{c}{5.5e-4}  & \multicolumn{1}{c}{4.3e-4} &  \multicolumn{1}{c}{8.1e-4} \\
\textbf{ours~(w/o TF)}  &  \multicolumn{1}{c}{1.1e-4}  &  \multicolumn{1}{c}{2.4e-4}        &   \multicolumn{1}{c}{6.2e-5}   & \multicolumn{1}{c}{8.9e-6} \\ 
\hline
\end{tabular}
}
\caption{We report the reconstruction error~($\downarrow$) over different materials for our method versus the multiple baselines.} 
\label{tbl:main_reconstruction}
\vspace{-1em}
\end{table}

\subsection{Inferring Physical Properties from 3D Trajectories} 
\begin{figure*}[!t]
  \centering
  \includegraphics[trim=0cm 0cm 11cm 0cm, clip, width=\linewidth]{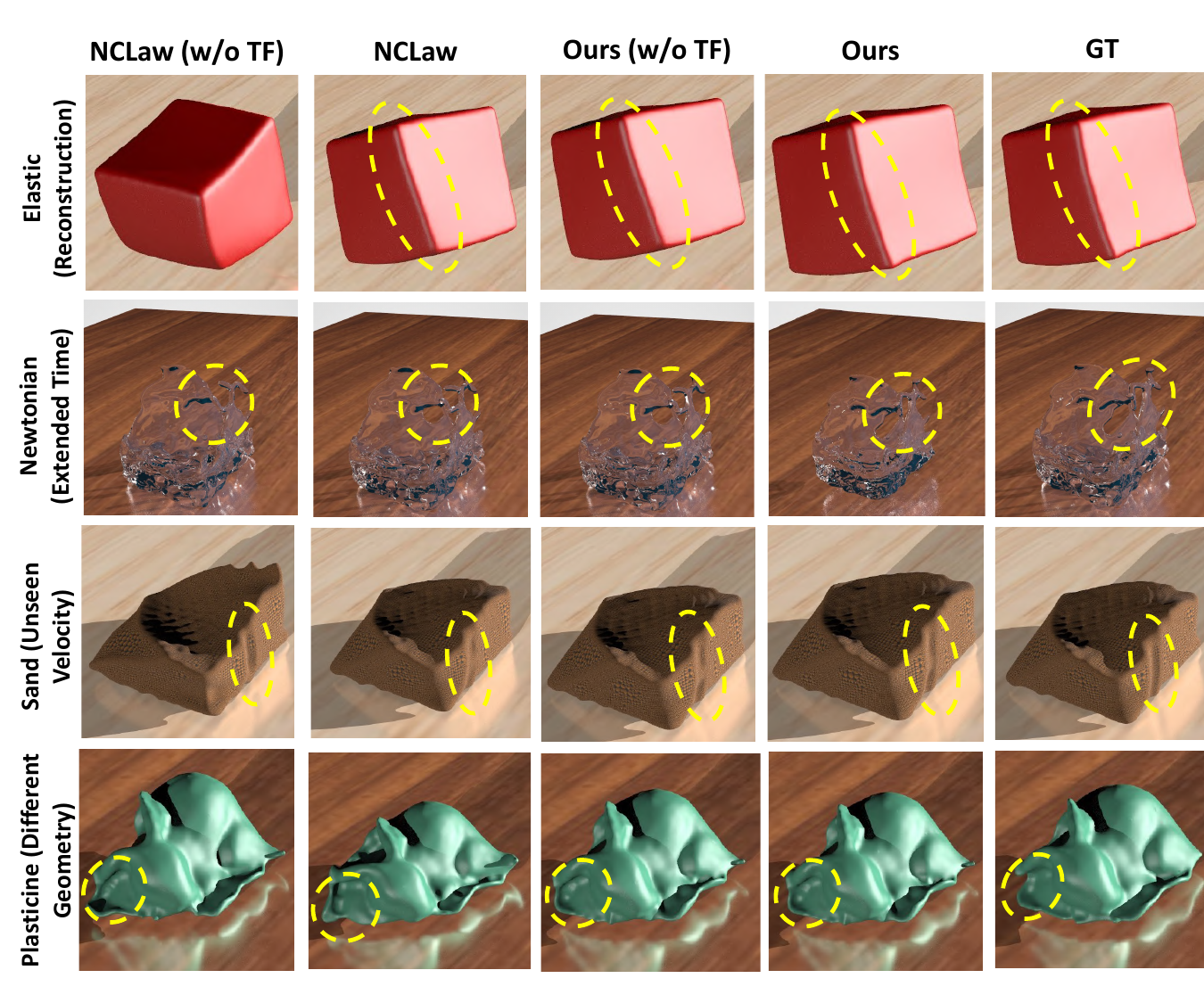}
  \vspace{-2em}
  \caption{We show our results~(w/ and w/o teacher forcing~(TF)) in columns four and three respectively compared with NCLaw~(w/ and w/o teacher forcing~(TF)) in columns two and one respectively, on reconstruction and generalization settings of extended time, unseen velocity, and different geometry. We show these differences in yellow dotted circles. We can see that in elastic our method is able to reconstruct the edge better. In newtonian, the separation between the fluid particles is ellipsoidal in ours and ground truth whereas the separation is circular in other visualizations. For sand, the fold of our method is more defined/sharper like GT as compared to the other visuals where the fold is smoother/softer. Finally, in plasticine, the spread and fold of the particles on the ground of ours and GT is similar.}
  \label{fig:qual_results}
\end{figure*}

We train \ours on our dataset to learn the latent space for all the materials. For inferring the material properties, we use the trajectory used by NCLaw~\cite{ma2023learning} and report the reconstruction error in Table~\ref{tbl:main_reconstruction}. We find that NCLaw consistently has a higher reconstruction error over all the materials than our method. This is likely because NCLaw is trained on a single trajectory of each material and fails to learn the diverse material properties. Since \ours learns a latent space across diverse materials for multiple trajectories, the rich representation of the latent space helps in developing a better understanding of the material properties and behavior. Similarly, we can also observe better reconstruction by our method~(fourth column) in Figure~\ref{fig:qual_results} where we are able to better reconstruct the edge of the jelly in comparison to NCLaw~(second column) which shows a more curved edge when compared to the ground truth. 

Additionally, NCLaw uses a teacher forcing scheme that periodically restarts the predicted simulation from the ground truth internal states. We believe that this introduces privileged information from the simulator which may not be available during inference if only the 3D positions are observed. Therefore, we also report a variant of ours and NCLaw without the teacher-forcing mechanism as Ours~(w/o TF) and NCLaw~(w/o TF) respectively and show that our method is able to perform better in this setting as well. In some cases, NCLaw~(w/o TF) is even unstable and unable to reconstruct well as seen in `elastic~(reconstruction)' example in Figure~\ref{fig:qual_results}.

\begin{table*}[h]
\resizebox{\linewidth}{!}{%
\centering
\begin{tabular}{l||ccc|ccc|ccc|ccc}
\hline
                   & \multicolumn{3}{c|}{\textbf{Elastic}}                                                                                       & \multicolumn{3}{c|}{\textbf{Sand}}                                                                                     & \multicolumn{3}{c|}{\textbf{Plasticine}}                                                                                    & \multicolumn{3}{c}{\textbf{Newtonian}}                                                                                         \\ \cline{2-13} 
                   & \multicolumn{1}{c}{\textbf{(a)}} & \multicolumn{1}{c}{\textbf{(b)}} & \multicolumn{1}{c|}{\textbf{(c)}} & \multicolumn{1}{c}{\textbf{(a)}} & \multicolumn{1}{c}{\textbf{(b)}} & \multicolumn{1}{c|}{\textbf{(c)}} & \multicolumn{1}{c}{\textbf{(a)}} & \multicolumn{1}{c}{\textbf{(b)}} & \multicolumn{1}{c|}{\textbf{(c)}} & \multicolumn{1}{c}{\textbf{(a)}} & \multicolumn{1}{c}{\textbf{(b)}} & \multicolumn{1}{c}{\textbf{(c)}} \\ \hline
\textbf{spline}~\cite{xu2015nonlinear} & \multicolumn{1}{c}{2.6e-1} & \multicolumn{1}{c}{2.4e-1} & \multicolumn{1}{c|}{3.5e-1} & \multicolumn{1}{c}{3.5e-1} & \multicolumn{1}{c}{3.1e-1} & \multicolumn{1}{c|}{3.6e-1} & \multicolumn{1}{c}{3.0e-1} & \multicolumn{1}{c}{3.0e-1} & \multicolumn{1}{c|}{3.1e-1} & \multicolumn{1}{c}{3.7e-1} & \multicolumn{1}{c}{3.2e-1} & \multicolumn{1}{c}{3.3e-1} \\ 
\textbf{neural}~\cite{wang2020learning} & \multicolumn{1}{c}{2.9e-5} & \multicolumn{1}{c}{1.4e-5} & \multicolumn{1}{c|}{5.6e-5} & \multicolumn{1}{c}{4.7e-2} & \multicolumn{1}{c}{6.2e-3} & \multicolumn{1}{c|}{2.3e-1} & \multicolumn{1}{c}{2.4e-1} & \multicolumn{1}{c}{4.1e-3} & \multicolumn{1}{c|}{2.8e-1} & \multicolumn{1}{c}{4.6e-2} & \multicolumn{1}{c}{1.7e-2} & \multicolumn{1}{c}{5.1e-2} \\
\textbf{gnn}~\cite{sanchez2020learning} & \multicolumn{1}{c}{3.3e-2} & \multicolumn{1}{c}{1.5e-2} & \multicolumn{1}{c|}{1.6e-1} & \multicolumn{1}{c}{1.5e-2} & \multicolumn{1}{c}{1.7e-2} & \multicolumn{1}{c|}{3.4e-1} & \multicolumn{1}{c}{1.1e-2} & \multicolumn{1}{c}{6.8e-3} & \multicolumn{1}{c|}{3.7e-2} & \multicolumn{1}{c}{1.1e+0} & \multicolumn{1}{c}{5.8e-2} & \multicolumn{1}{c}{2.9e-1} \\
\textbf{nclaw}~\cite{ma2023learning} & \multicolumn{1}{c}{9.8e-4} & \multicolumn{1}{c}{2.4e-4} & \multicolumn{1}{c|}{4.1e-4} & \multicolumn{1}{c}{4.2e-5} & \multicolumn{1}{c}{6.5e-5} & \multicolumn{1}{c|}{3.6e-4} & \multicolumn{1}{c}{1.4e-4} & \multicolumn{1}{c}{4.6e-5} & \multicolumn{1}{c|}{2.3e-4} & \multicolumn{1}{c}{3.5e-4} & \multicolumn{1}{c}{1.9e-5} & \multicolumn{1}{c}{2.4e-4} \\
\textbf{ours} & \multicolumn{1}{c}{\textbf{1.8e-5}}  & \multicolumn{1}{c}{\textbf{9.0e-6}}   & \multicolumn{1}{c|}{\textbf{4.4e-5}} & \textbf{3.0e-5}  & \textbf{5.2e-5} & \textbf{2.7e-4} & \multicolumn{1}{c}{\textbf{8.6e-5}}   &  \multicolumn{1}{c}{\textbf{3.4e-5}}  & \multicolumn{1}{c|}{\textbf{9.8e-5}} & \multicolumn{1}{c}{\textbf{2.6e-5}} & \multicolumn{1}{c}{\textbf{5.7e-6}} & \multicolumn{1}{c}{\textbf{2.8e-5}} \\ 
\hline
\textbf{nclaw~(w/o TF)} & 7.1e-3 & 5.5e-3 & 4.7e-3 & 3.9e-4 & 2.8e-4  & 6.1e-3 & 4.9e-3  & 2.2e-4 & 7.6e-3 & 8.0e-3  & 3.1e-4 & 5.1e-3 \\
\textbf{ours~(w/o TF)} &  7.5e-4 &  9.5e-5  &  1.2e-4  &  3.9e-5  & 6.0e-5  & 3.2e-4 &  9.7e-5 & 4.1e-5  &  1.3e-4  &  9.6e-5  & 8.8e-6  &   1.1e-4 \\

\hline                          
\end{tabular}
}
\caption{\textbf{Generalization Results}: We report the reconstruction error~($\downarrow$) over unseen settings of (a). extended time trajectory, (b). unseen velocity, (c). different geometry. In bold case, we show that our method is able to perform better than the baselines.} 
\label{tbl:generalization}
\vspace{-1em}
\end{table*}

\begin{table*}[h]
\centering
\resizebox{0.7\linewidth}{!}{%
\begin{tabular}{l||ccccc} 
\hline
                                          & \textbf{Elastic} & \textbf{Newtonian} & \textbf{Plasticine} & \textbf{Sand} & \textbf{non-Newtonian} \\ \hline
Random Latent: $g_\phi(., z')$ + $f_\theta(.,z')$                  &                  7.3e-5&                    1.4e-5&                     1.2e-4&               2.7e-4&                        9.1e-5\\
GT Deformation projection + $f_\theta$~(., z) &                  1.3e-8&                    5.2e-7&                     6.0e-8&               3.6e-6&                        1.1e-8\\
$g_\phi$~(., z) + GT Constitutive law               &                  9.4e-8&                    8.6e-7&                     8.8e-8&               7.2e-6&                        9.0e-8\\ 
Optimized Latent: $g_\phi(., z)$ + $f_\theta(.,z)$                   &                  2.2e-7&                    9.1e-7&                     1.9e-7&               8.5e-6&                       8.2e-7\\ \hline
\end{tabular}
}
\caption{\textbf{Latent analysis}: We report the reconstruction error and compare our trained latents of the material with their respective latent, latent of another trajectory, using ground truth deformation gradient projection function + $f_\theta$, and using $g_\phi$ with ground truth constitutive law and report the reconstruction error~($\downarrow$) over 10 trajectories from the training data.} 
\label{tbl:ablation}
\vspace{-1em}
\end{table*}

\begin{table}[h]
\vspace{-0.5em}
\small
\resizebox{\linewidth}{!}{%
\centering
\begin{tabular}{c||cccc}
\hline
\multicolumn{1}{c||}{\textbf{Latent Size}} & \multicolumn{1}{c}{\textbf{Elastic}} & \multicolumn{1}{c}{\textbf{Sand}} & \multicolumn{1}{c}{\textbf{Plasticine}} & \multicolumn{1}{c}{\textbf{Newtonian}} \\ \hline 
\multicolumn{1}{c||}{4} & \multicolumn{1}{c}{7.8e-4} & \multicolumn{1}{c}{3.4e-4}  & \multicolumn{1}{c}{1.7e-4} &  \multicolumn{1}{c}{2.9e-5} \\
\multicolumn{1}{c||}{32} & \multicolumn{1}{c}{1.1e-4} & \multicolumn{1}{c}{2.4e-4}  & \multicolumn{1}{c}{6.2e-5} &  \multicolumn{1}{c}{8.9e-6} \\
\multicolumn{1}{c||}{256} & \multicolumn{1}{c}{1.5e-4}  &  \multicolumn{1}{c}{2.7e-4} &   \multicolumn{1}{c}{7.1e-5}   & \multicolumn{1}{c}{9.4e-6} \\ 
\hline
\end{tabular}
}
\vspace{-1em}
\caption{Ablation of reconstruction error on the latent size} 
\label{tbl:latent_size}
\vspace{-1em}
\end{table}

\begin{figure}[!t]
  \centering
  \includegraphics[width=\linewidth]{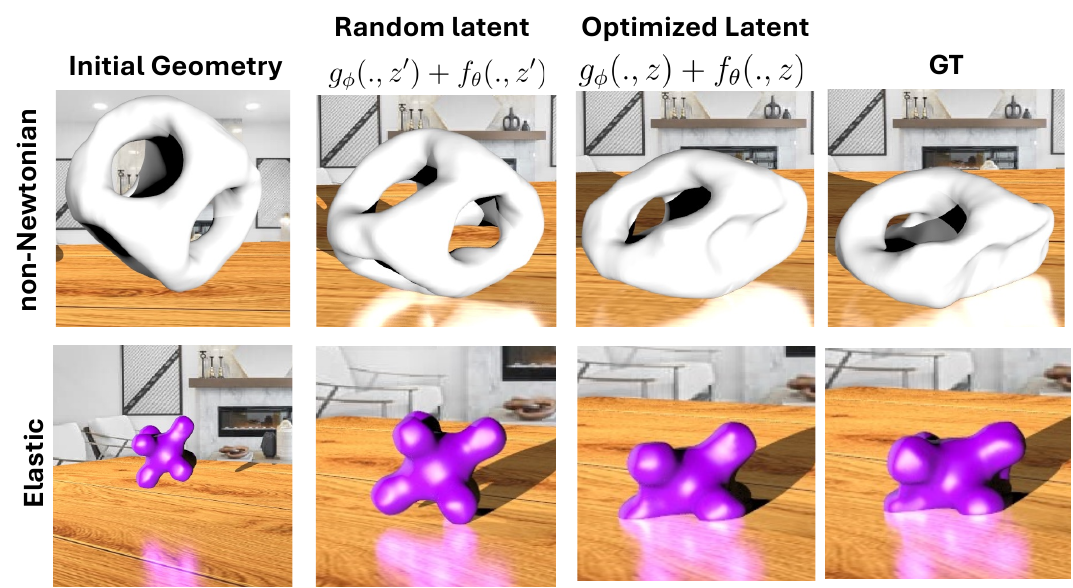}
  \caption{Qualitative Results on non-Newtonian and Elastic materials. The difference in the shape of the optimized latent~(ours) versus latent from other random trajectory shows that our method is able to learn the material properties of non-Newtonian materials~(as well as other materials) when compared with optimized versus other trajectory latent.}
  \label{fig:ablation_figure}
\end{figure}

\subsection{Ablations and Analysis}
\paragraph{Inferred material latent properties show generalization to scenario in novel configurations:} To show that the rich representation of the latent space of \ours is able to capture the material properties, we also show the generalization ability of our method on the tasks of extended time~(doubled time horizon), unseen velocity~(linear and angular velocity), and different geometry, and report the results in Table~\ref{tbl:generalization}. We can observe that over all the challenging/unseen scenarios, our method is able to generalize better than the baselines. We further demonstrate this generalization from the visualizations in the Figure~\ref{fig:qual_results} and show the visual differences in yellow dotted circle. For Newtonian fluids, we can see that the separation of the fluid particles for our method is closer to the separation in the ground truth which is ellipsoid in shape whereas the other methods show a circular separation. For sand, the trough/fold is more defined in our method versus the other methods which have softer/smoother fold. Finally, in plasticine, the spread and fold of the particles on the ground is more defined in our visualization when compared to other methods.

\paragraph{Specificity of Optimized Latents:} To validate that the optimization/inference of latents is crucial for accurate simulation, we conduct an ablation where we compare the simulated replay using our optimized latents vs a random latent optimized for a different trajectory. We show the quantitative results in Table~\ref{tbl:ablation} and compare our method's performance with the latent learned by the model for that trajectory during training~(optimized-latent) and the latent of a random trajectory in the training data~(random latent). The better performance of the optimized latent vs the random latent~(which can be of other materials) shows that the optimized latent is able to capture the material properties that are specific and unique. From the qualitative results in Figure~\ref{fig:ablation_figure}, when optimized latent and random latent random latent columns are compared, we observe that the optimized latent has a closer reconstruction with the ground truth versus the random latent which has a completely different geometry. We also remark that \ours\ can accurately model the properties of non-Newtonian fluids~\cite{yue2015continuum}, which has proven to be difficult for prior work~\cite{ma2023learning}.

\vspace{-1.3em}
\paragraph{Analysis of the combination of neural network and ground truth material function:} Our framework uses two neural networks for learning the constitutive law and the deformation gradient projection function conditioned on the latent. We perform an analysis where we keep one neural network and the other network is replaced with the ground truth material function and report the results in Table~\ref{tbl:ablation}. In both scenarios, the simulated trajectories are quite similar to the ground-truth, indicating that both networks have learned to accurately model the physical properties.

\vspace{-1em}
\paragraph{Latent size ablation:}
We perform an analysis of the reconstruction error on the size of the latent dimension and show the results in Table~\ref{tbl:latent_size}. We believe that a small latent dimension may not be expressive, whereas a larger latent can be hard to optimize.

\section{Discussion}
In this work, we present \ours, a latent-conditioned neural constitutive model that can capture the properties of diverse materials such as elastic, plasticine, sand, Newtonian and non-Newtonian fluids. Learning via latent helps us to infer the material properties without the assumption of knowing the material information beforehand. Better reconstruction results than the baselines across the diverse materials
on multiple generalization tasks shows that the latent is able to capture the material properties. 

While we showed that integrating \ours\ with the differentiable MPM simulator provides a promising mechanism for accurately inferring material properties from motion observations, there are several unaddressed limitations and challenges. First, our approach cannot handle non-homogeneous scenarios and it would be interesting to allow inference of multiple latents/materials in a scene. Moreover, our optimization required known initial geometry and (3D) motion observations, and these can be challenging to obtain. Nevertheless, we believe our framework of learning a unified constitutive model is a step towards the long-term goal of physical inference via inverse simulation `in-the-wild'.

\paragraph{Acknowledgments:} We thank the members of the Physical Perception Lab at CMU for their valuable
feedback. This work was supported by a Snap gift award and NSF Award IIS-2345610.

{
    \small
    \bibliographystyle{ieeenat_fullname}
    \bibliography{main}
}

\maketitlesupplementary

\section{Implementation Details}
\label{sec:rationale}

\textbf{Teacher Forcing}: The baseline, NCLaw~\cite{ma2023learning} uses a teacher-forcing scheme that restarts the predicted simulation from ground truth state periodically. The ground truth state includes the position $\inpx$, velocity $\inpv$, affine velocity $\inpC$, and deformation gradient $\inpF$. This period starts from 25 steps and is increased to 200 by a cosine annealing scheduler. This introduces the privileged information of position, velocity, affine velocity, and deformation gradient from the simulator during inference. Since the privileged information and access to the simulator may not be available during inference, we evaluate our method with the baseline NCLaw on the setting of without teacher-forcing and report the results in Table 1 and Table 2 of the main paper and show the visualizations in the webpage attached in the supplementary.

\section{Algorithm}
We detail our training and inference algorithm in Algorithm~\ref{algo:training} and Algorithm~\ref{algo:inference} respectively where $i$ is the trajectory index, $p$ represents the particle index and $t$ represents the time.

\begin{algorithm*}
	\caption{UniPhy: Training} 
	\begin{algorithmic}[1]
    
    \State \textbf{Input:} Dataset of trajectories $\Dcal = \{\mathbf{F}, \mathbf{F}_{proj}$, $\mathbf{C}, \mathbf{S} \}$
    \State \textbf{Output:} $\phi, \theta, z$
    \For {$iteration=1,2,\ldots$, N}
        \State Batch of $n$ samples, $\mathcal{B} = (\mathbf{F}^{p,t}_i, \mathbf{F}^{p, t}_{proj, i}, \mathbf{S}^{p, t}_{i}, \mathbf{C}^{p, t}_{i}, z_i)$ 
        \State $\mathbf{F}^{p,t}_i \stackrel{\text { SVD }}{=} \boldsymbol{U} 
        \boldsymbol{\Sigma} \boldsymbol{V}^T$
        \State $\Delta$ = $g_\phi$($\mathbf{F}^{p,t}_i$, $\boldsymbol{U}$, $\boldsymbol{V}^T$, $z_i$)
        \State $\mathbf{\hat{F}}^{p,t}_i$ = $\mathbf{F}^{p,t}_i$  + $\Delta$
        \State $\mathbf{F}$ = $\mathbf{F}^{p, t}_{proj, i}$
        \State $\mathbf{F}_{max}$ = $max(\mathbf{F}[:, 0, 0], 1e-6)$
        \State $\mathbf{F} \stackrel{\text { SVD }}{=} \boldsymbol{U}_{proj} \boldsymbol{\Sigma}_{proj} \boldsymbol{V}_{proj}$
        \State $\boldsymbol{R}=\boldsymbol{U}_{proj} \boldsymbol{V}_{proj}^T$
        \State $\boldsymbol{S}_1$ = $f_\theta\big(\boldsymbol{\Sigma}, 
        \mathbf{F}^{T} \mathbf{F}, \operatorname{det}(\boldsymbol{F}),
        log(\operatorname{det}(\boldsymbol{F})), 
        \mathbf{F}_{max}, log(\mathbf{F}_{max}), 
        \mathbf{C}^{p, t}_{i}, 
        z_i \big)$
        \State $\mathbf{\hat{S}}^{p, t}_{i} =\frac{1}{2}\left(\boldsymbol{S}_1+\boldsymbol{S}_1^T\right)$
        \State $\underset{\theta,\phi, {z}}{\min } \left(\mathcal{L} (\mathbf{\hat{F}}^{p,t}_i, \mathbf{F}^{p, t}_{proj, i}) + \mathcal{L}(\mathbf{\hat{S}}^{p, t}_{i}, \mathbf{S}^{p, t}_{i}) + \frac{1}{\sigma^2}\Vert z_i \Vert^2 \right)$
        \State Optimize $\phi, \theta, z$
    \EndFor
	\end{algorithmic} 
    \label{algo:training}
\end{algorithm*}

\begin{algorithm*}
	\caption{UniPhy: Inference using Differentiable Material Point Method~(MPM)} 
	\begin{algorithmic}[1]
    \State \textbf{Input:} $\mathbf{x}, z, f_\theta, g_\phi$
    \State \textbf{Output:} $\hat{z}$
    \For {$epoch=1,2,\ldots$,N}
        \For {$iteration=1,2,\ldots$,t}
            \State Transfer mass and momentum of particles to grid nodes
            \State $\mathbf{F}^{p, t+1} = (I + \Delta t * \mathbf{C}^{p, t}) * \mathbf{F}^{p, t}$
            \State $\mathbf{F}^{p,t+1} \stackrel{\text { SVD }}{=} \boldsymbol{U} 
            \boldsymbol{\Sigma} \boldsymbol{V}^T$
            \State $\Delta = g_\phi (\mathbf{F}^{p, t+1}, \mathbf{U}, \mathbf{V}^T, z_i)$
            \State $\mathbf{F}_{proj}^{p, t+1} = \mathbf{F}^{p, t+1} + \Delta$
            \State $\mathbf{F}_{proj}^{p, t+1} \stackrel{\text { SVD }}{=} \mathbf{U}_{proj} \mathbf{\Sigma}_{proj} \mathbf{V}_{proj}$
            \State $\boldsymbol{R}_{proj} =\mathbf{U}_{proj} \mathbf{V}_{proj}^T$
            \State $\boldsymbol{S}_1$ = $f_\theta\big(\boldsymbol{\Sigma}, 
            \mathbf{F}^{T} \mathbf{F}, \operatorname{det}(\boldsymbol{F}),
            log(\operatorname{det}(\boldsymbol{F})), 
            \mathbf{F}_{max}, log(\mathbf{F}_{max}), 
            \mathbf{C}^{p, t}_{i}, 
            z_i \big)$
            \State $\mathbf{\hat{S}}^{p, t}_{i}=\frac{1}{2}\left(\boldsymbol{S}_1+\boldsymbol{S}_1^T\right)$
            \State Update momentum and velocity of grid node
            \State Transfer momentum and velocity from grid node to particle
            \State Advect particles $\hat{\inpx}^{t+1} = \inpx^{t} + \Delta t \inpv^{t+1}$
        \EndFor
        \State $\hat{z} = \underset{{z}}{\min}~ \mathcal{L}\left(\hat{x}, {x} \right)$
        \State Optimize $\hat{z}$
    \EndFor
	\end{algorithmic} 
    \label{algo:inference}
\end{algorithm*}

\section{Analytical Constitutive Laws}

In this section, we discuss the constitutive model and the deformation gradient projection/mapping function for the materials that are used to simulate the trajectories used in training. 

In Material Point Method~(MPM), each particle has a deformation gradient $F$ which is projected on to the yield surface using a return mapping $\mathcal{G}$. This projected deformation gradient is then used by the constitutive law to compute the internal forces experienced by the particle given as the Cauchy stress $\mathcal{S}$.

\noindent \textbf{Elastic:} As there is no plasticity in elastic materials, the deformation gradient projection is an identity function defined as:

\begin{equation}
    \mathcal{G}(F) = F
\end{equation}

We use the neo-Hookean elasticity model for elastic materials. The Cauchy stress for the elastic material is calculated as:

\begin{equation}
    J\mathcal{S}(\mathbf{F})=\mu(\mathbf{F F}^{\top})+(\lambda \log (J)-\mu) \mathbf{I}
\end{equation}

where $\mu$ and $\lambda$ are the Lam\'e parameters of the Young's modulus and Poisson ratio. The Young's modulus defines the stiffness of the material and Poisson ratio defines the ability of the object to preserve its volume under deformation.

For elastic materials, we have a range of [350.0, 2595196.0] for $\mu$ and a range of [500.0, 2580120.0] for $\lambda$.

\textbf{Newtonian:} The stress for the newtonian fluid is computed as: 

\begin{equation}
    \kappa = \frac{2}{3}\mu + \lambda
\end{equation}

\begin{equation}
J\mathcal{S}(\mathbf{F})=\kappa\mathbf{I}(J-\frac{1}{J^6}) + \frac{1}{2} \mu\left(\nabla \mathbf{v}+\nabla \mathbf{v}^{\top}\right)
\end{equation}

where $\nabla \mathbf{v}$ is the affine velocity of the particle $\mathbf{C}$, $\mu$ represents the velocity change opposition and $\kappa$ is volume preservation ability.

For newtonian materials, we have a range of [50.0, 1e3] for $\mu$ and a range of [30.0, 5e5] for $\lambda$.

\textbf{Plasticine:} For plasticine materials, we use the von-Mises plastic return mapping for deformation gradient. The SVD of $\mathbf{F}$ can be defined as $F = U \Sigma V$ where $\epsilon = log(\Sigma)$ is the Hencky strain. The von Mises yield condition is defined as:

\begin{equation}
    \delta \gamma=\|\hat{\boldsymbol{\epsilon}}\|-\frac{\tau_Y}{2 \mu}
\end{equation}

where $\boldsymbol{\epsilon}$ is the normalized Hencky strain, ${\tau_Y}$ is the yield stress determining the plastic flow and the stress required for causing permanent deformation/yielding behavior. In the above yielding condition, if $\delta \gamma > 0$, then the deformation gradient breaks the yield constraint and is projected back into the elastic region via the following mapping:

\begin{equation*}
    \mathcal{G}(\mathbf{F})= \begin{cases}\mathbf{F} & \delta \gamma \leq 0 \\ 
    \mathbf{U} \exp \left(\boldsymbol{\epsilon}-\delta \gamma \frac{\hat{\epsilon}}{\|\hat{\epsilon}\|}\right) \mathbf{V}^{\top} & \delta \gamma > 0\end{cases}\\
\end{equation*}

To calculate stress, we use the St.Venant-Kirchhoff (StVK) constitutive model. 

\begin{equation}
    J \mathbf{S}(\mathbf{F})=\mathbf{U}(2 \mu \boldsymbol{\epsilon}+\lambda \operatorname{Tr}(\boldsymbol{\epsilon})) \mathbf{U}^{\top}
\end{equation}

For plasticine materials, we have a range of [1e4, 1e6] for $\mu$, a range of [1e4, 3e6] for $\lambda$ and a range of [5e3, 1e4] for ${\tau_Y}$.

\textbf{Sand:} To simulate sand particles, we use the Drucker-Prager~\cite{klar2016drucker} yield criteria as follows: 

\begin{equation}
    \operatorname{tr}(\boldsymbol{\epsilon})>0, \quad \text { or } \quad \delta \gamma=\|\hat{\boldsymbol{\epsilon}}\|_F+\alpha \frac{(3 \lambda+2 \mu) \operatorname{tr}(\boldsymbol{\epsilon})}{2 \mu}>0 . \\
\end{equation}

where $\alpha=\sqrt{\frac{2}{3}} \frac{2 \sin \theta_{\text {fric }}}{3-\sin \theta_{\text {fric }}}$ and $\theta_{\text {fric }}$ is the friction angle determining the slope of the sand pile. Then, we use the deformation gradient projection function as follows:

\begin{equation}
\mathcal{G}(\mathbf{F})= \begin{cases}\mathbf{U V}^{\top} & \operatorname{tr}(\boldsymbol{\epsilon})>0 \\ \mathbf{F} & \delta \gamma \leq 0  \&  \operatorname{tr}(\boldsymbol{\epsilon}) \leq 0 \\ \mathbf{U} \exp \left(\boldsymbol{\epsilon}-\delta \gamma \frac{\hat{\epsilon}}{\|\epsilon\|}\right) \mathbf{V}^{\top} & \delta \gamma > 0  \&  \operatorname{tr}(\boldsymbol{\epsilon}) \leq 0 \end{cases}
\end{equation}

For sand materials, we have a range of [2400.0, 9e6] for $\mu$, a range of [2400.0, 9e6] for $\lambda$ and a range of [0.01, 0.4] for $\theta_{\text {fric }}$.

\textbf{Non-newtonian:} To model non-newtonian materials~\cite{chhabra2006bubbles}, we use the viscoplastic model~\cite{yue2015continuum,fang2019silly} and von-Mises criteria to define the elastic region. Having the viscoplastic model prevents the deformation from being directly mapped back onto the yield surface. Non-newtonian materials have yield stress as well. We define the deformation gradient return mapping as follows:

\begin{align}
\hat{\mu}&=\frac{\mu}{d} \operatorname{Tr}\left(\boldsymbol{\Sigma}^2\right) \\
s&=2 \mu \hat{\epsilon} \\
\hat{s}&=\|\boldsymbol{s}\|-\frac{\delta \gamma}{1+\frac{\eta}{2 \hat{\mu}  \Delta t}} \\
\mathcal{Z}(\mathbf{F})&= \begin{cases}\mathbf{F} & \delta \gamma \leq 0 \\
\mathbf{U} \exp \left(\frac{\hat{s}}{2 \mu} \hat{\boldsymbol{\epsilon}}+\frac{1}{d} \operatorname{Tr}(\boldsymbol{\epsilon}) \mathbf{1}\right) \mathbf{V}^{\top} & \delta \gamma > 0 \end{cases}
\end{align}

For non-newtonian materials, we have a range of [1e3, 2e6] for $\mu$, a range of [1e3, 2e6] for $\lambda$, a range of [1e3, 2e6] for ${\tau_Y}$  and a range of [0.1, 100.0] for $\eta$.



\end{document}